\documentclass{article}

\usepackage{graphicx,epstopdf}
\usepackage[format=plain,
justification=raggedright,singlelinecheck=false]{caption}

\usepackage{color}
\usepackage[dvipsnames]{xcolor}

\usepackage{amssymb}

\usepackage[normalem]{ulem}

\usepackage{multirow}

\usepackage{siunitx}

\usepackage{booktabs}

\usepackage{arydshln}
\usepackage{times}
\usepackage{graphicx} % more modern
\usepackage{subfigure}

\usepackage{epstopdf}

\usepackage{natbib}

\usepackage{algorithm}
\usepackage{algorithmic}

\usepackage[colorlinks=true,citecolor=blue]{hyperref}

\usepackage[accepted]{icml2016}

\usepackage[english]{babel} 
\usepackage[T1]{fontenc}
\usepackage{times}

\usepackage{amsmath}

\makeatletter
\newcommand{\vast}{\bBigg@{3}}
\newcommand{\Vast}{\bBigg@{4}}
\makeatother

\usepackage{enumitem}

\icmltitlerunning{Hawkes Processes with Stochastic Excitations}

\begin{document}
	\setlength{\abovedisplayskip}{7pt} \setlength{\abovedisplayshortskip}{7pt}
	\setlength{\belowdisplayskip}{7pt} \setlength{\belowdisplayshortskip}{7pt}

	\newtheorem{theorem}{Theorem}
	\newtheorem{definition}{Definition}
	\newtheorem{remark}{Remark}
	\newtheorem{proposition}{Proposition}
	\newtheorem{corollary}{Corollary}
	\newtheorem{assumption}{Assumption}
	\newtheorem{lemma}{Lemma}
	\newcommand{\p}{\mathbb{P}}
	\newcommand{\q}{\mathbb{Q}}
	\newcommand{\e}{\mathbb{E}}

\twocolumn[
\icmltitle{Hawkes Processes with Stochastic Excitations}

\icmlauthor{Young Lee$^*$}{young.lee@nicta.com.au}
\vspace*{1.5pt}
%\icmladdress{Data61/NICTA, Canberra, Australia}
\icmlauthor{Kar Wai Lim$^\dagger$}{karwai.lim@anu.edu.au}
%\icmladdress{Data61/NICTA and ANU, Canberra, Australia}
\vspace*{1.5pt}
\icmlauthor{Cheng Soon Ong$^\dagger$}{chengsoon.ong@anu.edu.au}
\vspace*{1.5pt}
\icmladdress{$^*$Data61/National ICT Australia \& London School of Economics}%
\vspace*{-8pt}
\icmladdress{$^\dagger$Data61/National ICT Australia \& Australian National University}%
\icmlkeywords{self-exciting process, Hawkes process, stochastic differential equation, Metropolis Hastings, Gibbs sampling, exponential Langevin, geometric Brownian motion}

\vskip 0.5in]

\begin{abstract}
    We propose an extension to Hawkes processes by treating the levels of self-excitation as a stochastic differential equation. Our new point process allows better approximation in application domains where events and intensities accelerate each other with correlated levels of contagion. We generalize a recent algorithm for simulating draws from Hawkes processes whose levels of excitation are stochastic processes, and propose a hybrid Markov chain Monte Carlo approach for model fitting. Our sampling procedure scales linearly with the number of required events and does not require stationarity of the point process. A modular inference procedure consisting of a combination between Gibbs and Metropolis Hastings steps is put forward. We recover expectation maximization as a special case. Our general approach is illustrated for contagion following geometric Brownian motion and exponential Langevin dynamics.
\end{abstract}

\vspace{1mm}
\section{Introduction}

\vspace{1mm}
\paragraph{Motivation.} Cascading chain of events usually arise in nature or society: The economy has witnessed that financial meltdowns are often epidemic. For example, the Asian financial crisis swept across Thailand and quickly engulfed South Africa, Eastern Europe and even Brazil. Similarly, criminological research \cite{burglar} has shown that crime can spread through local environments very rapidly where burglars will constantly attack nearby targets because local susceptibilities are well known to thieves. As another example, in genetic analysis, \citet{bouret} looked at the likelihood of occurrences of a particular event along the DNA sequence where `an event' could be any biological signals occurring along the genomes that tend to cluster together.

The defining characteristic of these examples is that the occurrence of one event often triggers a series of similar events. The Hawkes process, or otherwise known as the self-exciting process, is an extension of Poisson processes that aims to explain excitatory interactions \cite{hawkes71}.
What makes the term \textit{self-excitation} worthy of its name is typically not the occurrence of the initial event, but the intensification of further events. We seek to characterize this amplification magnitude, which we call the \textit{contagion parameters}, or \textit{levels of self-excitation}, or simply, \textit{levels of excitation}.

The use of Hawkes processes is not an attempt to describe all features of self-excitation in their correct proportions. Probabilistic modeling inevitably exaggerates some aspects while disregarding others, and an accurate model is one that takes care of the significant aspects and abandons the less important details. Thus, there are often two streams in modeling which are fairly contradictory; on the one hand, the model ought to mimic excitatory relationships in a real world application, and this pulls toward specifying wide families of processes. On the other hand, the model should be manageable and tractable which pulls in the direction of identifying simpler processes in which inference and parameter estimations are feasible. 

Adopting the latter view of establishing simpler processes, we present a version of self-exciting processes that permits the levels of excitation to be modulated by a stochastic differential equation (SDE). SDEs are natural tools used to describe rate of change between the excitation levels. Put differently, we attach some indeterminacy to these quantities where we model them as random values over time, rather than being a constant as in the classical Hawkes setting \cite{hawkes71,Ozaki1979}. Our formulation implies that the contagion parameters are random processes thus inheriting tractable covariance structures, in contrast to the set-up initiated by \citet{BremaudMassoulie2002} and \citet{biao-aap}, where contagion levels are independent and identically distributed (\textit{iid}) random.

\paragraph{Contributions.}
We present a model that generalizes classical Hawkes and new insights on inference. Our noteworthy contributions are as follows: (1) We propose a fully Bayesian framework to model excitatory relationships where the contagion parameters are stochastic processes satisfying an SDE. This new feature enables the control of the varying contagion levels through periods of excitation. (2) With $n$ denoting the counts of events, we design a sampling procedure that scales with complexity $\mathcal{O}(n)$ compared to a na{\"i}ve implementation of Ogata's modified thinning algorithm \cite{Ogata1981} which needs $\mathcal{O}(n^2)$ steps. (3) A hybrid of MCMC algorithms that provide significant flexibility to do parameter estimation for our self-exciting model is presented. In addition, we describe how to construct two SDEs over periods of unequal lengths and introduce general procedures for inference. (4) We conclude by making explicit deductive connections to two related areas in machine learning; (i) the `E-step' of the expectation maximization (EM) algorithm for Hawkes processes \cite{veen-schoenberg,Ogata1981} and (ii) modeling the volatility clustering phenomenon.

\section{Our Model : Stochastic Hawkes}

\subsection{Review of Poisson and Classical Hawkes Processes} 

This section recapitulates some pieces of counting process theory needed in what follows. The \textit{Poisson process} is frequently used as a model for counting events occurring one at a time. Formally, the Poisson process with constant intensity $\lambda$ is a process $N=\{N_t:=N(t):t\geq 0\}$ taking values in $S=\{0,1,2,...\}$ such that: (a) $N_0=0;$ if $s<t,$ then $N_s\leq N_t$, (b) $\p(N_{t+h}=n+m\,|\,N_t=n)$ takes values $\lambda h+o(h)$ if $m=1$, $o(h)$ if $m>1$, and $1-\lambda h + o(h)$ if $m=0$ where $o(h)$ denotes any function $h$ that satisfies $o(h)/h\rightarrow 0$ as $h\rightarrow 0$. In addition, if $s<t$, the number $N_t-N_s$ of events in the interval $(s,t]$ is independent to the times of events during $(0,s]$. We speak of $N_t$ as the number of `arrivals' or `events' of the process by time $t$. However, events usually do not arrive in evenly spaced intervals but naturally arrive clustered in time. The \textit{Classical Hawkes process} aims at explaining such phenomenon. It is a point process $N$ whose intensity $\lambda_t$ depends on the path mirrored by the point process over time. Precisely, the point process is determined by $\lambda_t$ through the following relations: $\p(N_{t+h}=n+m\,|\,N_t=n)$ takes values $\lambda_t h+o(h)$ if $m=1$, $o(h)$ if $m>1$, and $1-\lambda_t h + o(h)$ if $m=0$, where $\lambda_t=c_0+\sum_{i:t>T_i}c_1\exp(-c_2(t-T_i))$ for positive constants $c_0,c_1$ and $c_2$.

\subsection{Proposed Model Specification and Interpretation}

%%%
\begin{figure}[tb!]
%	\vskip -0.55in
	\begin{center}
		\includegraphics[width=1.0\columnwidth]{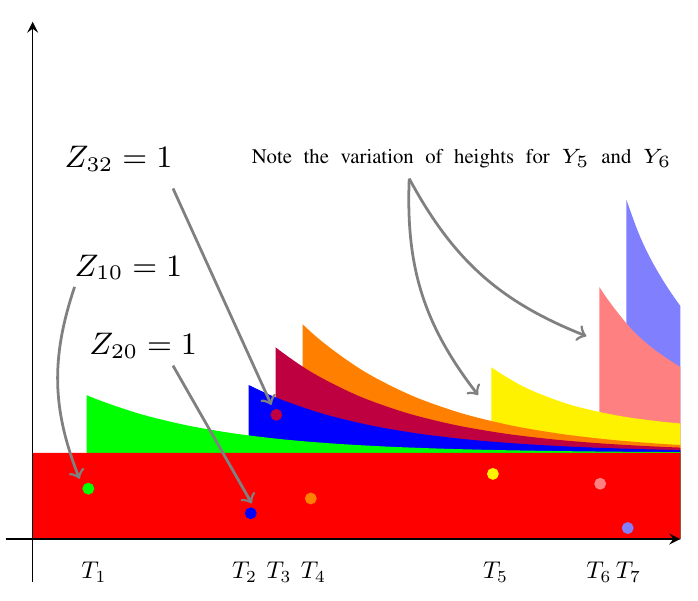}
		\caption{
			A sample path of the intensity function $\lambda(\cdot)$. First note that the \textcolor{red}{red} region represents the base intensity $\hat\lambda_0(t)$, which is assumed to be a constant in this diagram. Each colored region, except the \textcolor{red}{red}, represents the excitation contributed by each event time. For example, the \textcolor{blue}{blue} region is contributed by $T_2$, which in turn is represented by the \textcolor{blue}{blue} dot. Further, note that the \textcolor{blue}{blue} dot lies in the interior of the \textcolor{red}{red} region, indicating that it is an \textit{immigrant}. Mathematically, this is represented by $Z_{20}=1$. On the other hand, the offspring of the second event $T_2$ is represented by the \textcolor{purple}{maroon} dot, which is right on top of $T_3$. We denote this by $Z_{32}=1$. Observe that this offspring immediately induces another region to be conceived, which is consistently colored in \textcolor{purple}{maroon}. Stochastic Hawkes is capable of capturing and resembling different levels of contagion and this is evident from the differing heights in $Y$ at the event times $T_5$ and $T_6$, where we also allow for a non-zero covariance structure, i.e., $\mathrm{Cov}{(Y_5,Y_6)}\neq 0$. Finally, it is important to note that the higher the intensity $\lambda(\cdot)$ is, the stronger the rate of decay is. In other words, the gradient is the same for a fixed intensity level, which is a property of the exponential kernel $\nu$.
		}
		\label{fig:branching-structure}
	\end{center}
%	\vskip -0.2in
\end{figure}

%%%
We define our model as a linear self-exciting process $N(t)$ endowed with a non-negative $\mathcal{F}_t$--stochastic intensity function $\lambda(t)$:
\begin{align}
\label{first-lambda}
\lambda(t)=\hat{\lambda}_0(t)+\sum_{i:t>T_i}Y(T_i)\,\nu(t-T_i)
\end{align}
where $\hat{\lambda}_0:\mathbb{R}\mapsto\mathbb{R}_+$ is a deterministic base intensity, $Y$ is a stochastic process and $\nu:\mathbb{R}\mapsto\mathbb{R}_+$ conveys the positive influence of the past events $T_i$ on the current value of the intensity process. We write $N_t:=N(t), \lambda_t:=\lambda(t)$ and $Y_i:=Y(T_i)$ to ease notation and $\{\mathcal{F}_t\}$ being the history of the process and contains the list of times of events up to and including $t$, i.e. $\{T_1,T_2,...,T_{N_t}\}$.
Figure~\ref{fig:branching-structure} illustrates the different components of our model, which we
explain in the following subsections.

\subsubsection{Base Intensity, $\hat{\lambda}_0$}
\vspace{-1mm}
This parameter is the base or background intensity describing the arrival of external-originating events in the absence of the influence of any previous events. These events are also known as \textit{exogenous} events. By way of analogy, the base rate is referred to as the `immigrant intensity' in ecological applications \cite{Law}, where it describes the rate with which new organisms are expected to arrive from other territories and colonies. In our case $\hat{\lambda}_0(t)$ is a function of time and takes the form $\hat{\lambda}_0(t)=a+(\lambda_0-a)e^{-\delta t}$ where $\lambda_0>0$ is the initial intensity jump at time $t=0$, $a>0$ is the constant parameter, and $\delta>0$ is the constant rate of exponential decay.

\vspace{-1mm}
\subsubsection{The Contagion Process, $(Y_i)_{i=1,2,...}$}
\label{sec:contagion-process}

The levels of excitation $Y$ measure the impact of clustering or contagion of the event times. To see this, observe in Equation~(\ref{first-lambda}) that whenever $Y$ is high and of positive value, it imposes a greater value to the intensity $\lambda$, thus increasing the probability of generating an event in a shorter period of time, thereby causing the clustering phenomena.

We use differential equations to describe the evolution of the levels of excitation. Translating the evolution of contagiousness into the language of mathematics means setting up an equation containing a derivative (or an integral), discussed further below. The changes in the contagion is assumed to satisfy the stochastic differential equation
\begin{equation*}
Y_{\cdot} = \int_0^{\cdot}\hat{\mu}(t,Y_t) \, dt + \int_0^{\cdot}\hat{\sigma}(t,Y_t) \, dB_t
\end{equation*}
where $B$ is a standard Brownian motion and $t\in[0,T]$ where $T<\infty$. Different settings of the functionals $\hat{\mu}$ and $\hat{\sigma}$ lead to different versions of SDEs.

An important criterion for selecting appropriate choices of the couple $(\hat{\mu},\hat{\sigma})$ essentially boils down to how we decide to model the levels of excitation within Stochastic Hawkes. A standing assumption is that the contagion process has to be positive, that is,

%\vspace{-1.3em}
\vspace{-1mm}
\begin{assumption}
\label{ass:y-positive}
The contagion parameters $Y_t>0,\forall t\geq 0$.
\end{assumption}
\vspace{-1mm}
%\vspace{-0.6em}

This is necessary as the levels of excitations $Y$ act as a parameter that scales the magnitude of the influence of each past event and subsequently contributes to the quantity $\lambda$ in Equation~(\ref{first-lambda}), which is non-negative.

Some notable examples of the couple $(\hat{\mu},\hat{\sigma})$ are the Geometric Brownian Motion (GBM): $\hat{\mu}=(\mu+\frac{1}{2}\sigma^2) Y,\,\hat{\sigma}=\sigma Y$ \cite{kloeden-platen,zammit-pnas}; the Square-Root-Processes: $\hat{\mu}=k(\mu-Y),\,\hat{\sigma}=\sigma\sqrt{Y}$, \cite{archambeau,opper-sde};  Langevin equation: $\hat{\mu}=k(\mu-Y),\,\hat{\sigma}=\sigma$, and their variants \cite{stimberg-negative,welling,liptser-shiryaev}.

Whilst the positivity of $Y$ is guaranteed for GBM, this may not be true for other candidates such as the Langevin dynamics or the Square-Root-Processes. This is because they possess the inherent property that nothing prevents them from going negative and thus may not be suitable choices to model the levels of excitation. Specifically, Square-Root-Processes can be negative if the Feller condition $2k\mu>\sigma^2$ is not satisfied \cite{feller-condition,liptser-shiryaev}. For real-life applications, this condition may not be respected, thus violating Assumption~\ref{ass:y-positive}.

To that end, we focus on two specifications of the SDEs, namely the GBM and we tilt the Langevin dynamics by exponentiating it so that the positivity of $Y$ is ensured \cite{black-kara}:
\begin{itemize}[nosep]
	\item Geometric Brownian Motion (GBM):
	\begin{equation}
	\label{gbm}
	Y_{\cdot} = \int_0^{\cdot}\left(\mu+\frac{1}{2}\sigma^2\right) Y_t \, dt + \int_0^{\cdot}\sigma Y_t \, dB_t
	\end{equation}
	
	where $\mu\in\mathbb{R}$ and $\sigma>0$.
	\vspace*{3mm}
	
	\item Exponential Langevin:
	\begin{equation}
	\label{ou}
	Y_{\cdot} = \exp\left(\int_0^{\cdot}k(\mu-Y_t) \, dt + \int_0^{\cdot}\sigma \, dB_t\right)
	\end{equation}
	
	where $k,\mu\in\mathbb{R}$ and $\sigma>0$.
\end{itemize}

The parameter $k$ for exponential Langevin denotes the decay or growth rate and it signifies how strongly the levels of excitation reacts to being pulled toward the asymptotic mean, $\mu$. For fixed $\sigma$ and $\mu$, a small value of $k$ implies $Y$ is not oscillating about the mean. 

%%%%% full stretch 3 in a row, with 2 rows
\begin{figure*}[t!]
	\begin{center}
		\hspace*{-2.0em}
		\subfigure[\footnotesize{True contagion: GBM }]{\includegraphics[width=0.77\columnwidth]{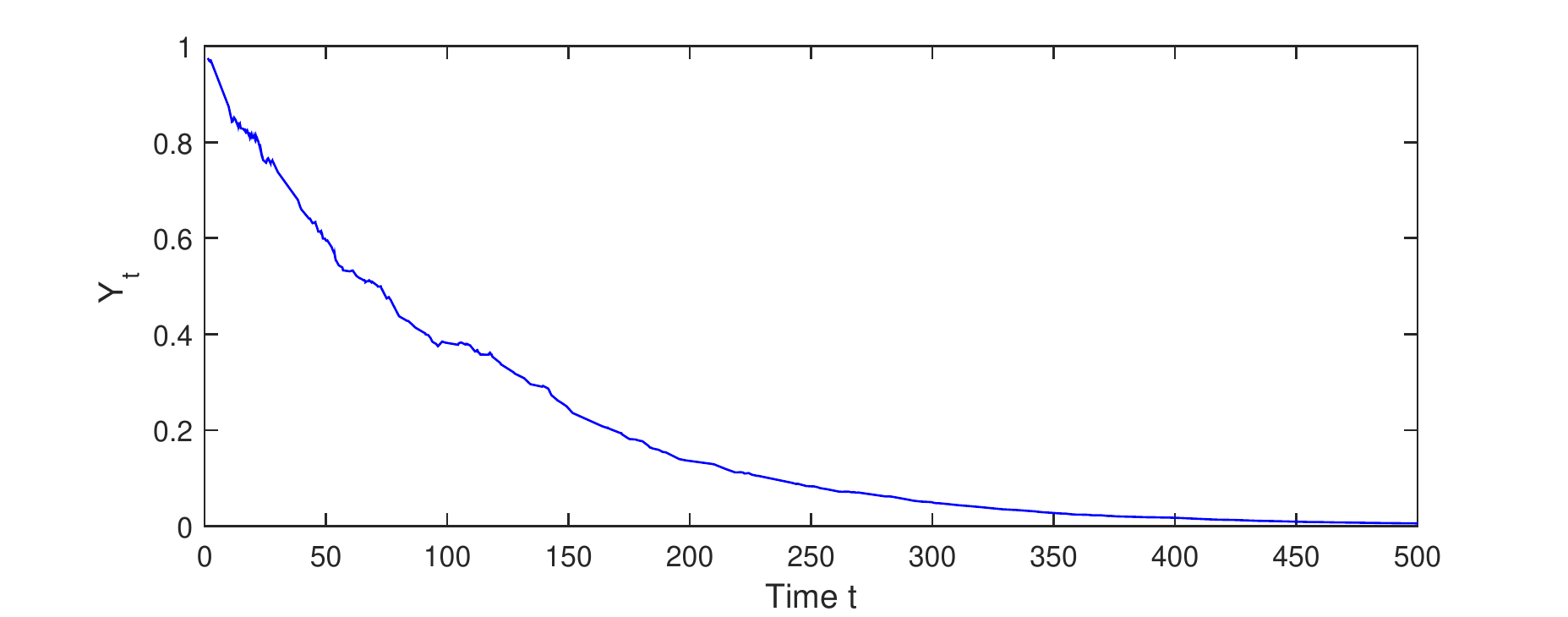}}\hspace*{-1.5em}
		\subfigure[\footnotesize{Contagion learned with GBM}]{\includegraphics[width=0.77\columnwidth]{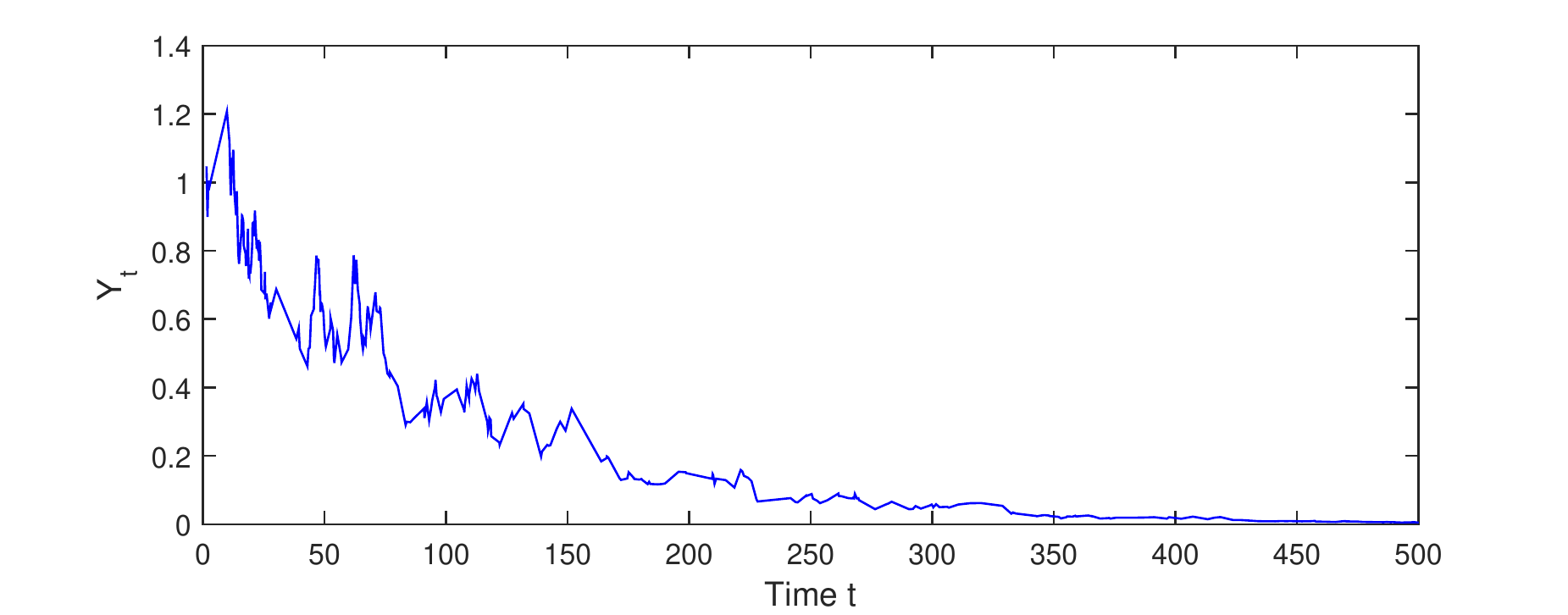}} \hspace*{-1.5em}
		\subfigure[\footnotesize{Contagion learned with Gamma}]{\includegraphics[width=0.77\columnwidth]{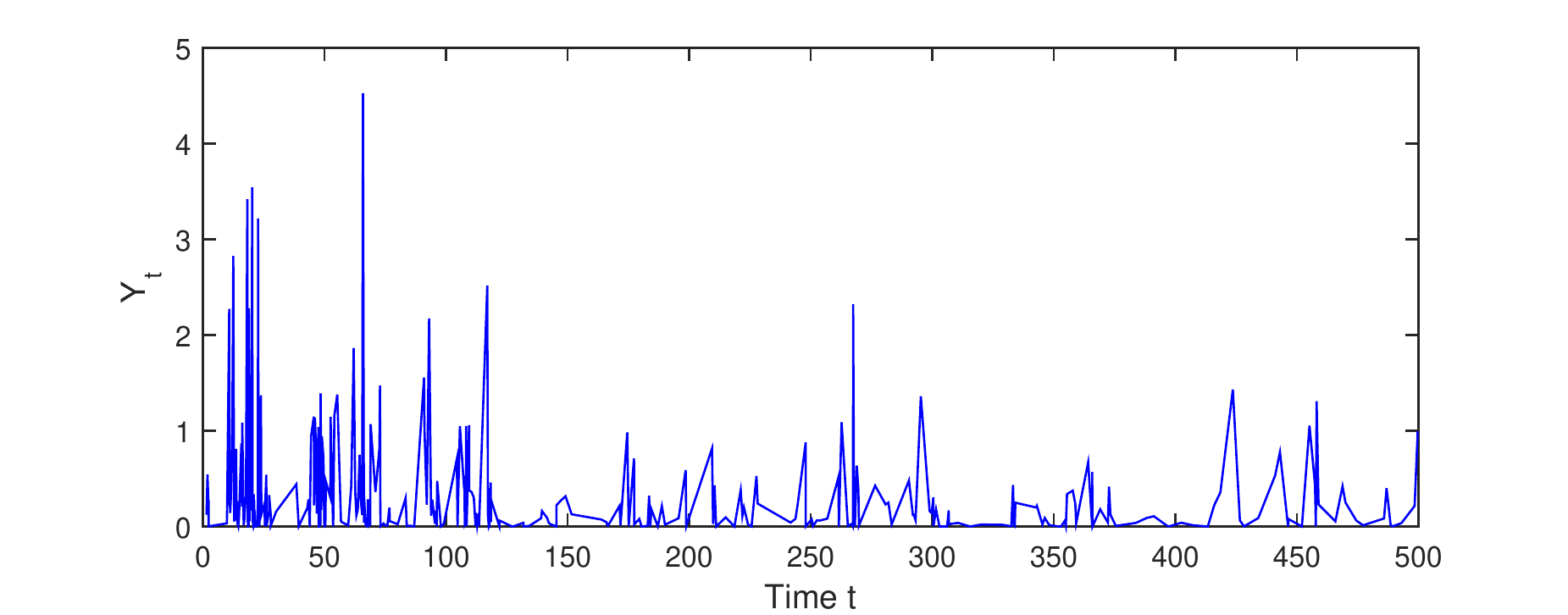}}\hspace*{-1.5em}
		
		\vspace*{-1.0em}
		\hspace*{-2.0em}
		\subfigure[\footnotesize{True contagion: Gamma}]{\includegraphics[width=0.77\columnwidth]{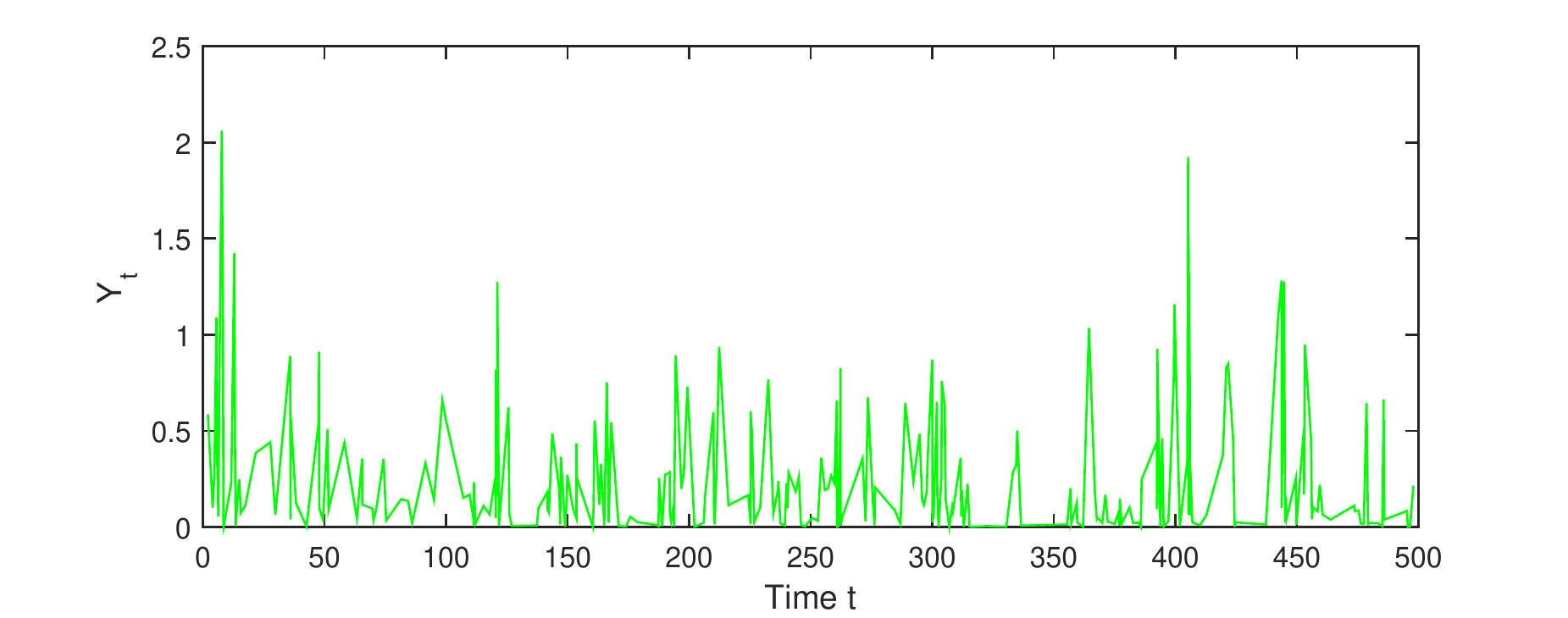}}\hspace*{-1.5em}
		\subfigure[\footnotesize{Contagion learned with Gamma}]{\includegraphics[width=0.77\columnwidth]{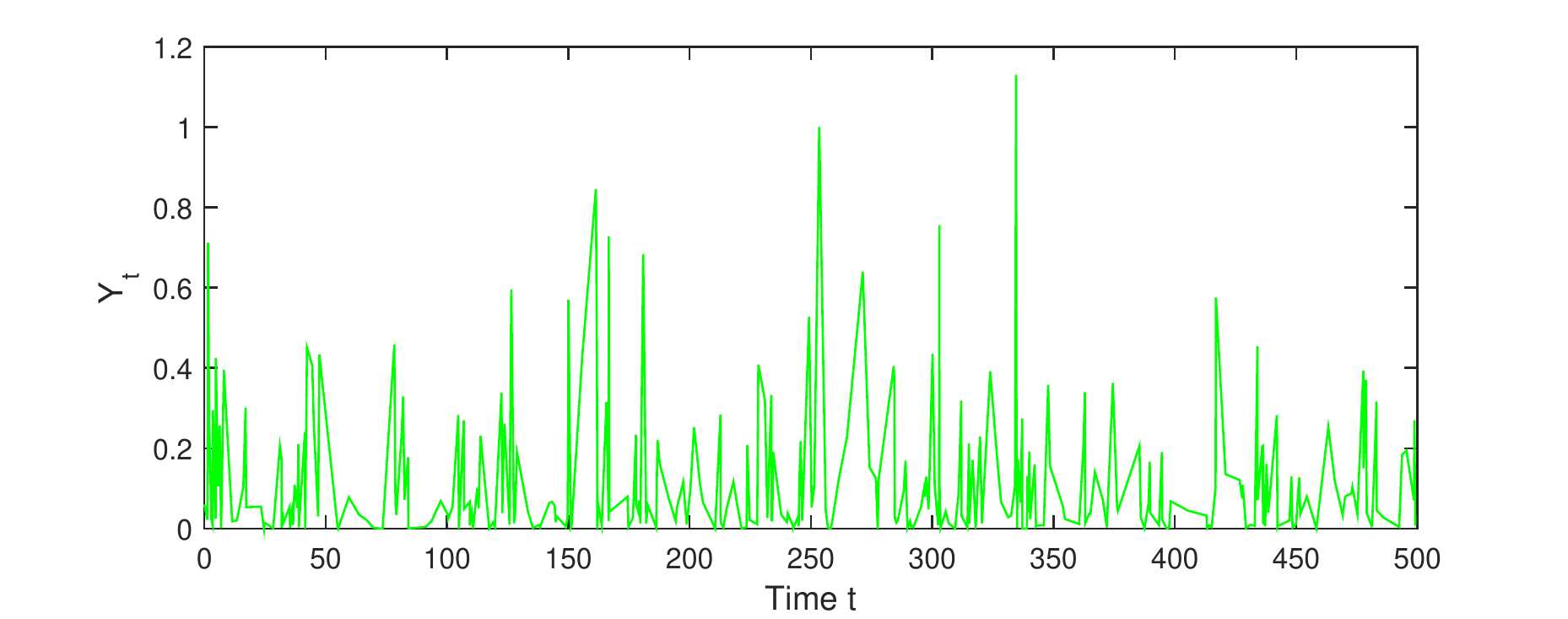}} \hspace*{-1.5em}
		\subfigure[\footnotesize{Contagion learned with GBM}]{\includegraphics[width=0.77\columnwidth]{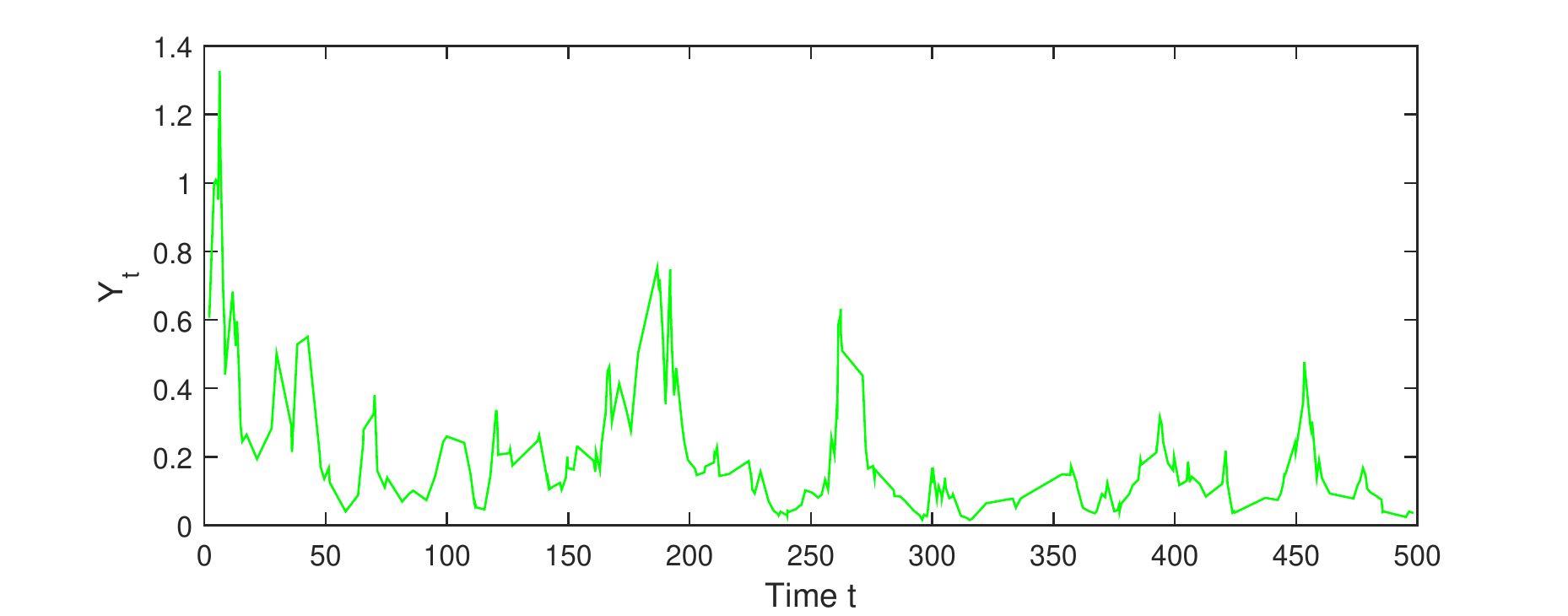}}\hspace*{-1.5em}
		\caption{Versatility of Stochastic Hawkes: Observe that by allowing the level of contagion $Y$ to be a stochastic process satisfying for instance a Geometric Brownian Motion (GBM) makes it possible to reproduce stylized facts of \textit{both} ground truths of GBM and \textit{iid} Gamma variates, see plots (a) \& (b) and (d) \& (f). On the other hand, it is not possible to perform inference of a more general class of stochastic process, if we were to start with \textit{iid} variables. For example, with the parametrization that $Y$ is \textit{iid} Gamma, one can only reproduce the ground truth when it follows \textit{iid} Gamma but not the stylized facts inherited by a GBM as seen from plots (a) \& (c). This can be seen by observing that the path of $Y$ in plot (c) is does not resemble that of the path $Y$ in plot (a).}
%		\vspace{-3.5mm}
		\label{fig:versatility}
	\end{center}
\end{figure*}

\vspace{-1mm}
\subsubsection{The Sum Product, $\sum_{i:t>T_i}Y(T_i)\,\nu(t\!-\!T_i)$. }
The product $Y\nu$ describes the impact on the current intensity of a previous event that took place at time $T_i$. We take $\nu$ to be the exponential kernel of the form $\nu(t)=e^{-\delta t}$. Note that $\delta$ is being shared between the base intensity $\hat\lambda_0$ and the kernel $\nu$ to ensure that the process has memoryless properties \citep[see][]{hawkes-oakes-cluster,Ozaki1979}. The memoryless property states the following: given the present, the future is independent of the past. We exploit this property to design our sampling algorithm so that we do not need to track all of its past history, only the present event time matters. This would not have been possible, if we were to use a power kernel \citep{ogata-power}, say. In addition, choosing this kernel enables us to derive Gibbs sampling procedures to facilitate efficient inference.

Summarizing, the intensity of our model becomes:
\begin{align}
\lambda_t
& = a + (\lambda_0-a)e^{-\delta t}
+ \sum_{i: \, T_i < t}^{N_t} Y_i \, e^{-\delta (t - T_i)}.
\label{eq:intensity_function}
\end{align}
If we set $Y$ to be a constant, we retrieve the model proposed by \citet{hawkes71}. In addition, setting $Y=0$ returns us the inhomogeneous Poisson process. Furthermore, letting $Y=0$ and $\lambda_0=a$ simplifies it to the Poisson process.

\subsection{The Branching Structure for Stochastic Hawkes}
This section presents a generative view of our model that permits a systematic treatment of situations where each of the observed event times can be separated into \textit{immigrants} and \textit{offsprings}, terminologies that we shall define shortly. This is integral to deriving efficient inference algorithms. 

We call an event time $T_i$ an \textit{immigrant} if it is generated from the base intensity $a + (\lambda_0-a) e^{-\delta t}$, otherwise, we say $T_i$ is an \textit{offspring}. This is known as the \textit{branching structure}. It is therefore natural to introduce a variable that describes the specific process to which each event time $T_i$ corresponds to. We do that by introducing the random variables $Z_{ij}$, where $Z_{i0}=1$ if event $i$ is an immigrant, and $Z_{ij}=1$ if event $i$ is an offspring of event $j$.
% An equivalent interpretation of the set $\{Z_{ij}=1\}$ is the following: event $i$ was caused by event $j$. Furthermore, each $Z_i$ is a special indicator matrix where only one of its element is unity, i.e., the vector $Z_i=(Z_{i0},Z_{i1},.....,Z_{i,(i-1)})$ contains a single 1 and 0 otherwise. For a fixed $i$, we have $\sum_{j=0}^{i-1} Z_{ij} = 1$. 
We illustrate and elucidate the branching structure through Figure~\ref{fig:branching-structure}.
For further details on classification of Hawkes processes via the branching structure, 
refer to \citet{rasmussen-bayesian} and \citet{DaleyBook}.

\subsection{Advantages of Stochastic Contagion}
Before we dive into the technical definitions of simulating samples for our point process and performing parameter inference, we illustrate the effect of model choice.
The levels of contagion $Y$ can necessarily come in different flavors: be it a constant, as in the standard classical Hawkes, or a sequence of \textit{iid} random variables, or satisfying an SDE.

First we generate levels of contagion $Y$ following a GBM, see Figure~\ref{fig:versatility}(a). Performing inference by learning $Y$ as a GBM leads to good estimation as indicated in Figure~\ref{fig:versatility}(b). However, if we let $Y$ to be Gamma distributed, we are less able to reproduce the properties of the ground truth, as is evident from Figure~\ref{fig:versatility}(c). This is fairly intuitive as the Gamma distribution does not inherit any serial correlation between the samples as they are \textit{iid}, whereas the ground truth does possess correlation structure of a Geometric Brownian Motion.

Proceeding further, this time with the ground truth $Y$ inheriting \textit{iid} Gamma random variables, as illustrated in Figure~\ref{fig:versatility}(d). Learning $Y$ as Gamma in our model leads to good inference as illustrated in Figure~\ref{fig:versatility}(e). Further, letting $Y$ to be a GBM enables us to learn some properties of the ground truth this time round, see Figure~\ref{fig:versatility}(f). Comparing Figures~\ref{fig:versatility}(a) \& (c) against Figures~\ref{fig:versatility}(d) \& (f), we conclude that a fairly general formulation for the level of contagion, such as the GBM, is advantageous in recovering stylized facts of an \textit{iid} random levels of contagion, but not vice versa.

\subsection{Likelihood Function}
This section explicates the \textit{likelihood} with the presence of an SDE and the branching structure within Stochastic Hawkes. The derivation is new and merits discussion here. A key result is that the integrated intensity function $\Lambda_t:=\int_0^t \lambda_v \, dv$ can be derived explicitly, that is,
\begin{align*}
&\int_0^t \lambda_v \, dv
=\int_0^t \left(a\!+\!(\lambda_0\!-\!a)e^{-\delta v}\! +\! \sum_{i=1}^{N_v} Y_i e^{-\delta(t-v)}\right)dv
\end{align*}
\begin{align*}
&\overset{(i)}{=}\int_0^t a+(\lambda_0-a)e^{-\delta v}dv+\int_0^t\int_0^v Y_s e^{-\delta(v-s)} dN_s dv
\end{align*}
\begin{align*}
&\overset{(ii)}{=}\int_0^t a+(\lambda_0-a)e^{-\delta v}dv+\int_0^t\int_s^t Y_s e^{-\delta(v-s)}dv dN_s
\end{align*}
\begin{align*}
&\overset{(iii)}{=}at + \frac{(\lambda-a)(1-e^{-\delta t})}{\delta}+ \frac{1}{\delta}\sum_{i=1}^{N_t}Y_i \Big( 1-e^{-\delta(t-T_i)} \Big)
\end{align*}
where we note the second term for $(i)$ is a Riemann integral over a stochastic integral with respect to $N$, $(ii)$ is due to Fubini's Theorem and $(iii)$ follows from the equivalence between stochastic integral and the sum over event times. Consequently, we get the following:
\begin{proposition}
	\label{proposition-likelihood}
	Let $\mathcal{T},\mathcal{Z},\mathcal{Y}$ be $\{T_i\}$,$\{Z_i\}$ and $\{Y_i\}$ for $i=1,2,...,N_T$ respectively. Assume that no events have occurred before time 0. Further set $\mathcal{J}_i:=\{j:0<j<i\}$. Then the likelihood function $\p(\mathcal{T}\,|\,\mathcal{Z},\mathcal{Y})$ is given by
	\begin{align*}
e^{-\Lambda_T} \prod\limits_{i=1}^{N_T} \big( a+(\lambda_0-a)e^{-\delta t} \big)^{Z_{i0}}\prod\limits_{j\in\mathcal{J}_i} \left[Y_j e^{-\delta(T_i-T_j)}\right]^{Z_{ij}}.
	\end{align*}
\end{proposition}

This generalizes the likelihood function of \citet{lewis-mohler}. It can be viewed that this as an alternative version of the likelihood function found in \citet{DaleyBook} and \citet{rubin} with the presence of the branching structure coupled with a stochastic process $Y$. The proof of this result can be found in the Supplemental Materials (Section B).

\section{Simulation of Stochastic Hawkes}
\label{sec:simulation}

\begin{algorithm}[t!]
	\caption{Simulation of Stochastic Hawkes}
	\label{alg:simulation-stochastic-hawkes}
	\begin{enumerate}[nosep]
		\item
		We firstly set $T_0 = 0$, $\lambda_{0}^{(1)} = \lambda_0-a$, and given $Y_0$.

		\item
		For $i = 1,2,\dots$ and while $T_i < T$:
		\begin{enumerate}[nosep]
			\item
			Draw $S_i^{(0)} = - \frac{1}{a} \log U(0,1)$.

			\item
			Draw $u \sim U(0,1)$. Set $S_i^{(1)} =
			- \frac{1}{\delta} \log
			\Big(
			1 - \delta/\lambda_{T_{i\!-\!1}}^{(1)}
			\log u
			\Big)$. Note we set $S_k^{(1)}:= \infty$ when the $\log$ term is undefined.
			\item Set $T_i = T_{i\!-\!1} +  \min \Big( S_i^{(0)}, S_i^{(1)} \Big)$.
			%\item Sample $Y_{T_k}$ {(\scriptsize{refer to Algorithm \ref{alg:appendix} in %Supplementary Materials})}
			\item Sample $Y_{T_i}$ {(\scriptsize{refer to Algorithm 1 in Supplemental Materials})}
			\item Update 
			$\lambda_{T_i}^{(1)}
			= \lambda_{T_{i\!-\!1}}^{(1)} \, e^{-\delta (T_i - T_{i\!-\!1})} + Y_{T_i}$.
		\end{enumerate}
	\end{enumerate}
\end{algorithm}

We present a sampling procedure for our Stochastic Hawkes model 
in Algorithm~\ref{alg:simulation-stochastic-hawkes}. 
This algorithm scales~with complexity $\mathcal{O}(n)$ compared to a na\"ive implementation of Ogata's modified thinning algorithm \cite{Ogata1981} which requires $\mathcal{O}(n^2)$ steps with $n$ denoting the number of events. Similarly to \citet{Ozaki1979} but differently from Ogata's method, it is noteworthy to mention that our algorithm does not require the stationarity condition for intensity dynamics as long as $T<\infty$.

The outline of \citet{biao-ecp} for simulating Hawkes processes is followed closely and adapted to the present setting. The idea of their algorithm is to decompose the inter-arrival event times into two independent simpler random variables, denoted by $S^{(0)}$ and $S^{(1)}$, with the intention that they can be sampled conveniently. Note however that we also need to sample the levels of self-excitation $Y$, which is a stochastic process in contrast to \textit{iid} sequences of $Y$ as in \citet{biao-ecp}.

We seek to find laws that describe the GBM and exponential Langevin dynamics. Applying It\^o's formula (see \citealp{liptser-shiryaev}, and Section C in the Supplemental Materials) on $f(y)=\log(y)$ and performing discretization for GBM yields
\begin{align}
\label{euler-maruyama}
Y_i = Y_{i\!-\!1}\exp\left(\mu \Delta_i +\sqrt{\sigma\Delta_i} \, \epsilon_i\right),
\end{align}
where $\Delta_i$ is introduced as a shorthand for $T_i - T_{i\!-\!1}$.
%Rearranging yields
%\begin{align}
%\log\left(\frac{Y_i}{Y_{i\!-\!1}}\right)&=\mu(T_i\!-\!T_{i\!-\!1})+\sqrt{\sigma(T_i\!-\!T_{i\!-\!1})}\epsilon_i.
%\\
%&\overset{\mathcal{D}}{=} N(\mu(T_i-T_{i\!-\!1}),\sigma^2(T_i-T_{i\!-\!1})).
%\end{align}
Similarly for exponential Langevin, the discretization scheme returns
\begin{align}
\nonumber
\log Y_i = (\log Y_{i-1}) \phi_i +\mu(1\!-\!\phi_i)
+ \sqrt{\frac{\sigma^2}{2k} \Big( 1-(\phi_i)^2 \Big)} \, \epsilon_i
\end{align}
where we define $\phi_i = e^{-k\Delta_i}$, $\epsilon_i\sim N(0,1)$ is standard normal, 
and $Y_0$ is known.
Both these expressions now allow us to sample $Y_i$ for all $i$. We state the following:

\begin{proposition}
	The simulation algorithm for a sample path of Stochastic Hawkes process is presented in Algorithm \ref{alg:simulation-stochastic-hawkes}.
\end{proposition}

The proof of this algorithm presented in the Supplemental Materials (Section A).

\section{Parameter Inference from Observed Data}
\label{sec:inference}
We present a hybrid of MCMC algorithms that updates the parameters one at a time, either by direct draws using Gibbs sampling or through the Metropolis--Hastings (MH) algorithm. A hybrid algorithm \cite{robert-casella} combines the features of the Gibbs sampler and the MH algorithm, thereby providing significant flexibility in designing the inference thereof for the parameters within our model.

To see the mechanics of this, consider a two-dimensional parameterization as an illustration. Let $\theta_A$ and $\theta_B$ be parameters of interest. Assume that the posterior $\p(\theta_B\,|\,\theta_A)$ is of a known distribution, we can perform inference directly utilizing the Gibbs sampler. On the other hand, suppose $\p(\theta_A\,|\,\theta_B)$ can only be evaluated but not directly sampled; then, we resort to the use of an MH algorithm to update $\theta_A$ given $\theta_B$. 
%For the MH step, the candidate $\theta^{\prime}_A$ is drawn from $\q(\theta^{\prime}_A\,|\,\theta^{(j)}_A,\theta^{(j)}_B)$, which indicates that the current step can depend on the past draw of $\theta_A$. 
The MH step samples from a proposal distribution $\q\big(\theta_A^{\prime}\,|\,\theta^{(j)}_A,\theta^{(j)}_B\big)$ which implies that we draw $\theta^{(j+1)}_A\sim \q\big(\theta^{\prime}_A\,|\,\theta^{(j)}_A,\theta^{(j)}_B\big)$ and that the criteria to accept or reject the proposal candidate is based on the acceptance probability, denoted by $AP\big(\theta^{(j+1)}_A\big)$:
\begin{equation}
\min\left(1,
\frac{
\p \big(\theta^{\prime}_A\,|\,\theta^{(j)}_B \big)\,\q \big(\theta^{(j)}_A\,|\,\theta^{\prime}_A,\theta^{(j)}_B \big)
}{
\p\big(\theta_A^{(j)}\,|\,\theta^{(j)}_B\big)\,\q\big(\theta^{\prime}_A\,|\,\theta^{(j)}_A,\theta^{(j)}_B\big)
}\right).
\label{eq:ap}
\end{equation}
The hybrid algorithm is as follows: given $\left(\theta^{(0)}_A,\theta^{(0)}_B\right)$, for $j=0,1,....,J$ iterations:
\begin{enumerate}[nosep]
	\item Sample $\theta_A^{(j+1)}\!\sim \q\big(\theta_A^{\prime}\,|\,\theta^{(j)}_A,\theta^{(j)}_B\big)$ and
	\textit{accept} or \textit{reject} $\theta_A^{(j+1)}$ based on Equation~(\ref{eq:ap}).
	\item Sample $\theta_B^{(j+1)}\!\sim \p\big(\theta_B\,|\,\theta^{(j+1)}_A\big)$ with Gibbs sampling.
\end{enumerate}
\vspace*{1mm}
We proceed by explaining the inference of a simple motivating example in Section \ref{sec:y-gamma}. This is the case when the contagion parameters $Y$ are \textit{iid} random elements. The main inference procedures for $Y$ being stochastic processes can be found in Sections~\ref{sec:gibbs-y-stochastic} and \ref{sec:mh-y-stochastic}. 

We summarize our MCMC algorithm in Algorithm \ref{alg:entire-mcmc}.

\subsection{Example: Levels of Excitation $Y$ are \textit{iid} Random}
\label{sec:y-gamma}
The focus here is on $Y$ to be \textit{iid} random elements with distribution function $G(y),y>0$. To form a suitable model for the problem under consideration, we propose to model $Y$ as a sequence of \textit{iid} Gamma distribution. This is a slight generalization to the Exponential distribution suggested by \citet{rasmussen-bayesian} as Gamma distribution contains an additional shape parameter that will help to improve the fitting performance.

The $Y_i$ are assumed to inherit \textit{iid} 
Gamma distribution with shape $\tau$ and scale $\omega$: 
$\p(y\,|\,\tau,\omega)\propto y^{\tau}e^{-\omega y}$. 
We also fix Gamma priors for $\{a,\lambda_0,\delta,\tau,\omega\}$ with hyperparameters $\{(\alpha_m,\beta_m)\text{ where }m=a,\lambda_0,\delta,\tau,\omega)\}$. Since all branching structure is equally likely \textit{a priori},
we have $\p(\mathcal{Z}) \propto 1$. The posterior
for $Y_i$ follows Gamma distribution, 
$
Y_i\,| \,
\cdot \! \sim \!
\Gamma
\big(
\tau\!+\! \sum_{r = i+1}^{N_T} Z_{ri} \ , \
\omega \!+\! \frac{1 - e^{-\delta (T\! -\! T_i)}}{\delta}
\big)
$
which can easily be sampled. Turning to the parameters of $Y$, we note that Gamma prior on  $\omega$ gives Gamma posterior,
$
\omega \,|\, \cdot \!
\sim \! \Gamma
\big(
\alpha_{\omega} + \tau N_T  , \
\beta_{\omega} + \sum_{i=1}^{N_T} Y_i
\big)
$
and the acceptance probability for the sampled $\tau'$
is given by $\min(1, A(\tau'))$ where $A(\tau')$ takes the form
$
\big(
\omega
^{N_T} \,
\prod_{i=1}^{N_T}
Y_i
\big)^{\tau' \!-\! \tau}
\big(
\frac{\tau'}{\tau}
\big)^{\alpha_{\tau} \!-\! 1}
\big(
\frac{\Gamma ( \tau' )}{\Gamma(\tau)}
\big)^{-N_T}
e^{
-
( \tau' \!-\! \tau )
\beta_{\tau}
}
$.

\begin{algorithm}[t!]
	\caption{MCMC Algorithm For Stochastic Hawkes}
	\label{alg:entire-mcmc}
	\begin{enumerate}[itemindent=0pt, itemsep=-0.5pt]
		\item
		Initialize the model parameters by sampling from their priors.
		\item For all $Z_i$: Use Gibbs sampler to generate a sequence of $Z_i$ using the posterior distribution defined in Equation~(\ref{eq:Zprobabilities}) with parameters derived in Section \ref{sec:samplingZ}.
		\item Depending on the choice of SDE so that for all $Y_i$: sample $Y_i$ using an MH scheme as tabulated in Table~1 in Supplemental Materials (Section D). %\ref{table:mh-entire},
		\item For $a$, $\lambda_0$ and $\delta$: sample these quantities with an MH scheme as tabulated in Table 1 in Supplemental Materials (Section D). %\ref{table:mh-entire},
		\item For the contagion parameters $\mu$, $\sigma^2$ and $k$, perform Gibbs sampling using the posterior parameters derived in Section \ref{sec:sampling-sde}.
		\item
		Repeat steps 2--6 until the model parameters converge or when a fixed number of iterations is reached.
	\end{enumerate}
\end{algorithm}
%\vspace*{-3mm}

\subsection{Gibbs Sampling}
\label{sec:gibbs-y-stochastic}

\subsubsection{Sampling the Branching Structure $\mathcal{Z}$}
\label{sec:samplingZ}
The posterior of $\mathcal{Z}$ follows the Multinomial posterior distribution
$
Z_i \,|\,\mathcal{T}, \mathcal{Y},\delta, a, \lambda_0
\sim
\mathrm{Multinomial}(\mu_{i\cdot})
\label{eq:samplingZ}
$
where $\mu_{i\cdot}=\mu_{ij}$ for all $j$ is a probability matrix (each row sum to 1) satisfying
\begin{align}
\label{eq:Zprobabilities}
\hspace{-2.6mm}
\mu_{ij}:=
\left\{
\begin{array}{ll}
\p(Z_{i0}=1)=\frac{a + (\lambda_0-a)e^{-\delta T_i}}{W_i}
& \hspace{-2.1mm} \mathrm{if \ }j = 0
\\[2mm]
\p(Z_{ij}=1)=\frac{Y_je^{-\delta (T_i-T_j)}}{W_i}
& \hspace{-2.1mm} \mathrm{if }\ 0 < j < i \!\!\!\!\!
\end{array}
\right.
\end{align}
where
\begin{align}
\label{eq:Znormalizer}
W_i&= a + (\lambda_0-a)e^{-\delta T_i}
+ \sum_{0<j<i} Y_j \, e^{-\delta (T_i-T_j)}
\end{align}
is a normalizing constant. In the Gibbs sampler, we sample new $Z_i$ directly from its posterior.

\subsubsection{Sampling the Contagion Parameters}
\label{sec:sampling-sde}

\paragraph{Geometric Brownian Motion.}

Let $X_i= \log(Y_i/Y_{i\!-\!1})$ and $\mathcal{X}=(X_1,X_2,...,X_{N_T})$. Given that the joint posterior is given by $\p(\mu,\sigma^2 \,|\, \mathcal{X})$, we take two independent conjugate priors, $\p(\mu)\sim N(\mu_0,\sigma_0^2)$ and $\p(\sigma^2)\sim\Gamma_{\mathrm{Inv}}(\alpha_0,\beta_0)$ where $\Gamma_{\mathrm{Inv}}$ refers to the inverse Gamma distribution. Standard calculations yield the posterior distributions
$\p(\mu\,|\,\sigma^2,\mathcal{X})\sim N(\mu_*,\sigma^2_*)$ and 
$\p(\sigma^2\,|\,\mu,\mathcal{X})\sim \Gamma_{\mathrm{Inv}}(\alpha_*,\beta_*)$
where the posterior parameters are given by
\begin{align}
&\!\!\!\!\,\; \mu_*=\frac{\sigma_0^2\sum_{i=1}^{N_T}X_i+\mu_0\sigma^2}{\sigma_0^2\sum_{i=1}^{N_T}\Delta_i+\sigma^2}, \ 
\sigma^2_*=\left(\frac{\sum_{i=1}^{N_T}\Delta_i}{\sigma^2}\!+\!\frac{1}{\sigma^2_0}\right)^{-1}
\!\!,
\end{align}
and
\begin{align}
&\!\!\!\!\,\; \alpha_*=\alpha_0 + \frac{N_T}{2}, \ \ 
\beta_* = \beta_0 + \frac{1}{2}\sum_{i=1}^{N_T}\frac{(X_i-\mu\Delta_i)^2}{\Delta_i}.
\end{align}

\vspace{-5mm}
\paragraph{Exponential Langevin.}

We take similar priors for $\mu,\sigma^2$ as in the case for GBM. We further assume that $k\sim N(\mu_k,\sigma^2_k)$. The posterior distributions for $\mu,\sigma^2$ and $k$ are $N(\hat\mu_*,\hat\sigma^2_*)$, $\Gamma_{\mathrm{Inv}}(\hat\alpha_*,\hat\beta_*)$ and $N(\hat\mu_k,\hat\sigma^2_k)$ with
\begin{align}
\hat\mu_*&\!=\!\frac{\sigma_0^2\sum_{i=1}^{N_T}(\log Y_i\! -\! \phi_i \log Y_{i-1})\xi_i\!+\!\mu_0\sigma^2}{\sigma_0^2\sum_{i=1}^{N_T}\xi_i\phi_i^-\!+\!\sigma^2},
\end{align}
as well as
\begin{align}
\hat\sigma^2_* &=\left(\frac{\sum_{k=1}^{N_T}\xi_i\phi_i^-}{\sigma^2}\!+\!\frac{1}{\sigma^2_0}\right)^{-1},\\[1mm]
\hat\alpha_*&=\alpha_0 \!+\! \frac{N_T}{2},
\end{align}
and also 
\begin{align}
\hat\beta_* = \beta_0\! +\! \frac{1}{2}\sum_{k=1}^{N_T}\frac{k(\log Y_i\! -\! \phi_i \log Y_{i\!-\!1}\!-\!\mu \phi_i^-)^2}{\phi_i^- \phi_i^+}.
\end{align}

Recall that the parameter $k$ expresses the \textit{wildness} of fluctuation about the mean level $\mu$. A small value of $k$ translates to a volatile $Y$. If we believe that the levels of self-excitations were erratic, which is of particular interest, then we would want a small value for $k$. This implies that expanding the power series on the exponential function $e^x\approx 1+x$ where $x:=-2k\Delta_i$ to the first order would be sufficient. This is similar in spirit to the Milstein scheme \citep{kloeden-platen} where higher orders of quadratic variations vanish. For an exact sampling of $k$, one needs to resort to an MH scheme. In most applications, we can even set $k$ to be a constant and do not perform inference for it. Proceeding, we obtain

\vspace{-2mm}
\begin{align}
\hat\mu_k&=\frac{\sigma^2_k\sum_{i=1}^{N_T}(\log Y_{i\!-\!1}\!-\!\mu)+\sigma^2 k_0}{\sigma^2_k\sum\limits_{i=1}^{N_T}\Delta_i(\log Y_i \!-\!\mu)^2 + \sigma^2},\\[3mm]
\hat\sigma^2_k &=\left(\frac{\sum_{k=1}^{N_T}\Delta_i(\log Y_i \!-\!\mu)}{\sigma^2}\!+\!\frac{1}{\sigma^2_k}\right)^{-1},
\end{align}

\vspace{1mm}
where we have used the following shorthand $\Delta_i=T_i-T_{i-1}$,
$\phi_i = e^{-k\Delta_i}$, $\phi_i^- = 1 - \phi_i$, $\phi_i^+ = 1 + \phi_i$, and
$\xi_i\!=\!2k/\phi_i^+$
%, and $\Delta_i^{(2)}\!=\!2k \phi_i^- / \phi_i^+$
throughout the calculations.

\subsection{Metropolis-Hastings}
\label{sec:mh-y-stochastic}
For the case of $Y$ following the GBM, we propose a symmetric proposal for $Y_i$ with $g(Y^{\prime}_i\,|\,Y_i)\sim N(Y_i,\sigma_Y^2)$. The posterior of $\mathcal{Y}$ is $\p(\mathcal{Y}\,|\,\mathcal{T},\mathcal{Z},\delta,\mu,\sigma^2)$. The acceptance probability $AP$ for $Y_i^{\prime}$ is $AP(Y_i^\prime)=\min(1,A(Y_i^\prime))$ where
\begin{align*}
&A(Y_i^\prime)=\exp\vast[-\frac{1}{\delta}(Y_i^\prime-Y_i)(1-e^{-\delta(T-T_i)})\\
&-\frac{1}{2\sigma^2\Delta_i}\bigg\{\left(\log\left(\frac{Y_{i+1}}{Y^\prime_i}\right)-\mu\Delta_i\right)^2\\
&\quad\quad -\left(\log\left(\frac{Y_{i+1}}{Y_i}\right)-\mu\Delta_i\right)^2   \bigg\} \,\mathbb{I}_{\{T_{i+1}<T\}}  \\
&-\frac{1}{2\sigma^2\Delta_i}\bigg\{\left(\log\left(\frac{Y^{\prime}_i}{Y_{i-1}}\right)-\mu\Delta_i\right)^2 \\
&\quad\quad -\left(\log\left(\frac{Y_i}{Y_{i-1}}\right)-\mu\Delta_i\right)^2   \bigg\} \vast] 
\left(\frac{Y^{\prime}_i}{Y_i}\right)^{\sum_{r=i+1}^{N_T}Z_{ri}-1}
\end{align*}
where we have defined $T_{N_T+1}=\infty$ and $\sum_r Z_{ri}=0$ when $i=N_T$.

For the case of $a^{\prime}$ with symmetry normal proposal, the acceptance probability is $\min\big(1, A(a') \big)$,
where
\begin{align*}
A \big( a' \big)   
& = 
\left[
\prod_{i=1}^{N(T)} 
\Bigg(
\frac{
	a^{\prime} + (\lambda_0-a^{\prime})e^{-\delta T_i}
}{
a  + (\lambda_0-a) \, e^{-\delta T_i}
}
\Bigg)^{Z_{i0}}
\right]
\bigg(
\frac{a'}{a}
\bigg)^{\alpha_a - 1}
\nonumber
\\
& \ \ \ \ \ 
\times
\exp 
\Bigg(
\big( a' - a \big)
\bigg(
\frac{1}{\delta}
\big( 1 - e^{-\delta T} \big)
- T
- \beta_a
\bigg) 
\Bigg)
\end{align*}
For the inferences of the remaining parameters $\lambda_0,\delta$, and $Y_i$ for exponential Langevin, the acceptance probabilities are shown in Table 1 in Supplemental Materials.

\section{Discussion and Related Work}

\paragraph{Reduction to EM.} 

We show that careful selection of specific priors yields posterior probabilities that coincide with the distribution that is taken under the E-step in the EM (expectation--maximization) algorithm methodology launched by \citet{veen-schoenberg}. They utilized the branching structure as a strategy for obtaining the maximum likelihood estimates of a classical Hawkes process which has intensity as in Equation~(\ref{eq:intensity_function}) with $Y$ being a constant ($\psi$). As in their paper, we define the variables $u_i$ associated with the $i-$th event time $T_i$ as $u_i=j$ if event $i$ is caused by event $j$ and $u_i=i$ if the event $i$ is an immigrant event. The unobserved branching structure $u_i$ is treated as the missing data and used to construct an EM algorithm. The conditional expected value of the complete data log-likelihoood can be written as
\begin{align*}
\nonumber
& Q\big(\vartheta;\vartheta^{(q)}\big)
\!=\!\e\Big[\!\log(\mathrm{complete\ data\ likelihood})|\,\mathcal{F}_T,\vartheta^{(q)}\Big]\\
&\!=\!\e\Bigg[\sum_{i=1}^{N_T}\!\mathbb{I}_{\{u_i=i\}}\!\log(a\!+\!(\lambda_0\!-\!a)e^{-\delta t})\!-\!\!\int_{T_i}^{T}\!\nu^{\dagger}\!(s\!-\!T_i)\,ds\!\\
 &\qquad+\sum_{i=1}^{N_T}\sum_{j\neq i}\mathbb{I}_{\{u_i=j\}}\log\nu^{\dagger}(T_i\!-\!T_j)|\,\mathcal{F}_T,\vartheta^{(q)}\Bigg].
\end{align*}
The following probabilities are used to find an expression for the conditional expected complete data log-likelihood:
\begin{align*}
&\p(u_i=j\,|\,\mathcal{F}_{T_i},T_i)\\
&\!=\left\{
\begin{array}{ll}
\!\!\!
\frac{\nu^\dagger(T_i-T_j)}{a^{(q)}+(\lambda^{(q)}_0-a^{(q)})e^{-\delta^{(q)}T_i} + \sum_{j:T_j<T_i}\nu^{\dagger}(T_i-T_j\,|\,\vartheta^{(q)})}
\ \, \text{(a)}
\\[4mm]
\!\!\!
\frac{a^{(q)}+(\lambda^{(q)}_0-a^{(q)})e^{-\delta^{(q)}T_i}}{a^{(q)}+(\lambda^{(q)}_0-a^{(q)})e^{-\delta^{(q)}T_i} + \sum_{j:T_j<T_i}\nu^{\dagger}(T_i-T_j\,|\,\vartheta^{(q)})}
\ \, \text{(b)}
\end{array}
\right.
\end{align*}
taking value (a) when $0<j<i$ and value (b) when $j=i$.

The kernel used by \citet{veen-schoenberg} is $\nu^{\dagger}(t)=\psi e^{-\delta t}$. Observe that this coincides with Equations~(\ref{eq:Zprobabilities}) and~(\ref{eq:Znormalizer}) in Section \ref{sec:samplingZ} when $Y$ is set to a constant $\psi$. These probabilities are analogous to the probabilities used to perform the thinning in the modified simulation algorithm \citep[see][]{Ogata1981,farajtabar2014,ValeraGomezRodriguez2015}.

\paragraph{The Renewal Equation Governing $\mathbb{E}(\lambda_t)$.} 
We provide an expression for $\mathbb{E}(\lambda_t)$ when $Y$ follows an SDE:
\begin{align*}
\lambda_t
%& = a + (\lambda_0-a)e^{-\delta t}
%+ \sum_{i: \, T_i < t}^{N_t} Y_i \, e^{-\delta (t - T_i)}\\
%\nonumber
& \overset{1.}{=} a + (\lambda_0-a)e^{-\delta t} + \int_0^t Y_s e^{-\delta(t-s)} \, dN_s\\
\nonumber
& \overset{2.}{=} a + (\lambda_0-a)e^{-\delta t}+ \int_0^t Y_s e^{-\delta(t-s)}\lambda_s \, ds \\
&\quad\quad + \int_0^t Y_s e^{-\delta(t-s)} \, d\left(N_s-\int_0^s\lambda_u \, du\right)\\
\e[\lambda_t]&\overset{3.}{=}a + (\lambda_0-a)e^{-\delta t} + \int_0^t\e[ Y_s]\,\e[\lambda_s] e^{-\delta(t-s)} \, ds.
\end{align*} 
where 1. rewrites $\lambda$ as a stochastic integral, 2. follows from subtracting  and adding the mean value process of $\lambda$ and 3. propagating expectation through the equation.

\paragraph{Other Point Processes.}
\citet{simma} proposed an EM inference algorithm for Hawkes processes and applied to large social network datasets. Inspired by their latent variable set-up, we adapted some of their hidden variable formulation within the marked point process framework  into our fully Bayesian inference setting. We have leveraged ideas from previous work on self-exciting processes to consequently treating the levels of excitation as random processes. \citet{slinderman} introduced a multivariate point process combining self (Hawkes) and external (Cox) flavors to study latent networks in the data. These processes have also been proposed and applied in analyzing topic diffusion and user interactions \cite{rodriguez,yang-zha}. \citet{farajtabar2014} put forth a temporal point process model with one intensity being modulated by the other. Bounds of self exciting processes are also studied in \cite{HansenReynaudBouretRivoirard2015}. Differently from these, we breathe another dimension into Hawkes processes by modeling the contagion parameters as a stochastic differential equation equipped with general procedures for learning. This allows much more latitude in parameterizing the self-exciting processes as a basic building block before incorporating wider families of processes. 

Studies of inference for continuous SDEs have been launc\-hed by \citet{archambeau}. Later contributions, notably by \citet{ruttor-sde} and \citet{opper-sde}, dealt with SDEs with drift modulated by a memoryless Telegraph or Kac process that takes binary values as well as incorporating discontinuities in the finite variation terms. We remark that the inference for diffusion processes via expectation propagation has been pursued by \citet{botond-ep}. We add some new aspects to the existing theory by introducing Bayesian approaches for performing inference on two SDEs, namely the GBM and the exponential Langevin dynamics over periods of unequal lengths.

\section{Final Remarks}
%The conclusions were succinctly made known in the abstract
%and were underpinned in the previous sections, but it is
%worthwhile reiterating the focal points. 
We extended the Hawkes process by treating the magnitudes of self-excitation as random elements satisfying two versions of SDEs. These formulations allow the modeling of phenomena when the events and their intensities accelerate one another in a correlated fashion. 

Which stochastic differential equation should one choose? We presented two SDEs of unequal lengths in this work. The availability of other SDEs in the machine learning literature leaves us the modeling freedom to maneuver and adapt to each real-life application. Each scenario presents quite distinct specifics that require certain amount of impromptu and improvised inventiveness. Finally, a flexible hybrid MCMC algorithm is put forward and connexions to the EM algorithm is spelled out.
%Generalizing beyond the Hawkes' self-excitation property by incorporating an \textit{externally-excited} Cox dimension would be an exciting avenue for future work.

 \vspace{-1mm}
\section{Acknowledgments}
Young Lee wishes to thank Aditya K.\ Menon for inspiring discussions.

This work was undertaken at NICTA. NICTA is funded by the Australian Government through the Department of
Communications and the 
%ARC
Australian Research Council 
through the ICT
Centre of Excellence Program.

\newpage
\bibliography{bib/biblio}
\bibliographystyle{apa}

\end{document}